\documentclass{article}

\usepackage{titling}
\usepackage{arxiv}
\usepackage{listings}
\usepackage[utf8]{inputenc} 
\usepackage[T1]{fontenc}    
\usepackage{url}            
\usepackage{booktabs}       
\usepackage{amsfonts}       
\usepackage{nicefrac}       
\usepackage{microtype}      
\usepackage{lipsum}		
\usepackage{graphicx}
\usepackage{natbib}
\usepackage{doi}
\usepackage{amsmath}
\usepackage{placeins}
\usepackage{fancyhdr}
\usepackage{adjustbox}
\usepackage{tabularx}
\usepackage{subfig}
\usepackage{authblk}

\lstnewenvironment{codeverbatim}[1][]{%
  \lstset{
    basicstyle=\ttfamily,
    frame=tb,
    #1
  }%
}{}

\title{\textnormal{AgGym: An agricultural biotic stress simulation
environment for ultra-precision management planning}}

\author[1]{\textbf{Mahsa Khosravi}}
\author[2]{\textbf{Matthew Carroll}}
\author[3]{\textbf{Kai Liang Tan}}
\author[4]{\textbf{Liza Van der Laan}} 
\author[4]{\textbf{Joscif Raigne}}
\author[5]{\textbf{Daren S. Mueller}} 
\author[4]{\textbf{Arti Singh}} 
\author[3]{\textbf{Aditya Balu}}
\author[4]{\textbf{Baskar Ganapathysubramanian}}
\author[4]{\textbf{Asheesh Kumar Singh}\thanks{To whom correspondence should be addressed. E-mails: \texttt{soumiks@iastate.edu} \& \texttt{singhak@iastate.edu}}}
\author[3]{\textbf{Soumik Sarkar}\textsuperscript{*}}

\affil[1]{Department of Industrial and Manufacturing Systems Engineering, Iowa State University, Ames, Iowa, USA.} 
\affil[2]{Iowa Soybean Association, Iowa, USA.}
\affil[3]{Department of Mechanical Engineering, Iowa State University, Ames, Iowa, USA.}
\affil[4]{Department of Agronomy, Iowa State University, Ames, Iowa, USA.}
\affil[5]{Department of Plant Pathology, Entomology and Microbiology, Iowa State University, Ames, Iowa, USA.}

\affil[*]{Corresponding authors}

\renewcommand{\headeright}{}
\renewcommand{\undertitle}{}

\hypersetup{
pdftitle={A template for the arxiv style},
pdfsubject={q-bio.NC, q-bio.QM},
pdfauthor={David S.~Hippocampus, Elias D.~Striatum},
pdfkeywords={First keyword, Second keyword, More},
}

\begin{document}
\maketitle

\begin{abstract}
Agricultural production requires careful management of inputs such as fungicides, insecticides, and herbicides to ensure a successful crop that is high-yielding, profitable, and of superior seed quality. Current state-of-the-art field crop management relies on coarse-scale crop management strategies, where entire fields are sprayed with pest and disease-controlling chemicals, leading to increased cost and sub-optimal soil and crop management. To overcome these challenges and optimize crop production, we utilize machine learning tools within a virtual field environment to generate localized management plans for farmers to manage biotic threats while maximizing profits. Specifically, we present AgGym, a modular, crop and stress agnostic simulation framework to model the spread of biotic stresses in a field and estimate yield losses with and without chemical treatments. Our validation with real data shows that AgGym can be customized with limited data to simulate yield outcomes under various biotic stress conditions. We further demonstrate that deep reinforcement learning (RL) policies can be trained using AgGym for designing ultra-precise biotic stress mitigation strategies with potential to increase yield recovery with less chemicals and lower cost. Our proposed framework enables personalized decision support that can transform biotic stress management from being schedule based and reactive to opportunistic and prescriptive. We also release the AgGym software implementation as a community resource and invite experts to contribute to this open-sourced publicly available modular environment framework. The source code can be accessed at: \href{https://github.com/SCSLabISU/AgGym}{https://github.com/SCSLabISU/AgGym}.

\end{abstract}


\pagestyle{fancy}
\fancyhf{} 
\fancyhead[C]{} 
\fancyfoot[C]{\thepage} 
\fancyhead[L]{\textit{AgGym: An agricultural biotic stress simulation environment for ultra-precision management planning}} 
\section{Introduction}
Pests and diseases are a major threat to crop yields and consequentially to food security. On a global scale, pest and disease outbreaks result in an average of 10-30\% crop yield losses annually \cite{Savary2019-ur}. Climate change also poses an increased threat of crop pests and diseases, especially as increasing temperatures in temperate regions could potentially allow for greater reproduction, expansion of geographic distribution, and increased overwintering survival rates \cite{Skendzic2021-hh, Juroszek2015-ll}. Precision agriculture approaches provide a promising approach to localized and just-in-time mitigation, ensuring profitable and sustainable farming. However, it has not been used in the management of pests/diseases in the way that it has been used in weed management, soil fertility, and planting~\cite{vsarauskis2022variable}. 

Recently, the paradigm of cyber-agricultural systems (CAS) has emerged, utilizing multimodal sensing, AI and machine learning (ML), and intelligent actuation to enable real-time monitoring and precise control of pests and diseases~\cite{Sarkar2024-CAS, bhakta2019state}. For example, new advances in remote sensing have enabled prescriptive herbicide applications using Unpiloted Aerial Vehicles (UAVs) \cite{stroppiana2018early}, and real-time sensors have been used effectively for precision spraying with no difference in yield from conventionally managed fields \cite{zanin2022reduction}. A recent work developed an agricultural digital twin for mandarins for designing individualized management practices at an intra-orchard scale~\cite{kim2024agricultural}. Overall, CAS can optimize interventions, reduce excessive chemical use, and provide actionable insights for healthier crops, increased yields, and sustainable farming practices. However, for the success of CAS, we need better and pertinent sensing, modeling and reasoning, and actuation strategies in a scale-agnostic and time-sensitive manner. 

In this paper, we focus on the modeling and reasoning aspect in the context of biotic stresses (i.e., pests and diseases) and their mitigation. While it is difficult to develop one unified model that works equally well for insect pests and diseases, we leverage some of their common features to establish a base model that can be refined in a context-specific manner. 
The early days of plant disease models were empirical and based on simple rules and graphs that incorporated information from disease cycles and weather~\cite{stern1959integration}. Today, we have moved to more dynamic models that include spatial and temporal aspects of the disease triangle and cycle. This incorporates the dynamics of plant disease epidemics and modelling of crop losses, which includes the study of plant pathogens on crop growth, development, and performance~\cite{rossi2010modelling,horsfall1959plant,de2007disease, krause1975predictive, gonzalez2023plant}. Crop entomologists and pathologists have also extensively studied multiple crops and pests interactions, which is the basis of many pest growth and development models, as well as models that estimate the effect on crop yield. These studies have been the basis for establishing Integrated Pest Management (IPM) approaches to manage crop losses (the integrated control concept), \cite{radcliffe2009integrated}, management practices such as economic injury levels (EILs) \cite{catangui2009soybean} and relative risks of plant disease \cite{fall2018case}. However, standard crop models and simulation platforms (e.g., EPIC \cite{williams1989epic}, DSSAT \cite{jones2003dssat}, and APSIM \cite{keating2003overview}) typically do not account for biotic stresses, such as insect-pests and diseases. 

Recently, efforts have begun to bring crop models and disease models together, e.g., Triki et al, \cite{triki2023coupling} developed the Mediation Interface for Model Inner Coupling (MIMIC) that allows researchers to integrate plant disease models with crop growth models. Still, most of these models operate at the field scale rather than smaller areas within fields, thus overlooking the inherent spatial heterogeneity within production fields. To develop robust models that simulate impact and control of biotic stresses, it is essential to consider plant health status, relationships between host-pathogen-environment, disease cycles, pathogen development, and infection spread characteristics. These are non-trivial; and a monolithic modeling approach may not be a solution for all situations. Hence, there is a need to develop a modular and open-source simulator environment, which will enable the community to come together to build and personalize various models that work for different crops, production systems, pathogens, geographies, and climates. 

Our work aims to take the first step in filling this gap by developing a modular framework for simulating the development and spread of biotic stresses as well as estimating corresponding yield losses with and without chemical treatment. Specifically, we develop a simulator, called AgGym that simulates the effects of biotic stresses. This simulator also serves as an open-source Gym environment~\cite{brockman2016openai} that provides researchers with a versatile platform to explore and experiment with various management strategies under different environmental conditions using algorithmic approaches such as reinforcement learning (RL). In this regard, a few related works have emerged recently. For example, Overweg \textit{et al.} proposed CropGym~\cite{overweg2021cropgym} that optimizes fertilizer application to maximize crop yield returns, which used the python crop simulation environment to simulate the inputs in the model. Gautron \textit{et al.} converts the monolithic DSSAT crop model software to a gym-like python library for deep RL agent interactions. It facilitates daily engagements between an RL agent and the simulated crop setting within DSSAT and has been utilized to enhance nitrogen management ~\cite{wu2022optimizing}. All these works proposed environments for crop yield maximization \textit{via} fertilizer management policies~\cite{shaikh2022machine, tonnang2017advances}. However, there is still a lack of validated Gym environments for designing biotic stress management strategies at a localized level and our work aims to bridge this gap. 
Specifically, we implement multiple popular deep RL algorithms within the AgGym environment and provide their baseline performances in optimizing chemical treatment schedules for real agricultural use cases. 
\section{Materials and Methods}
\label{sec:headings}
\begin{figure}
	\centering
	\includegraphics[width=.8\linewidth]{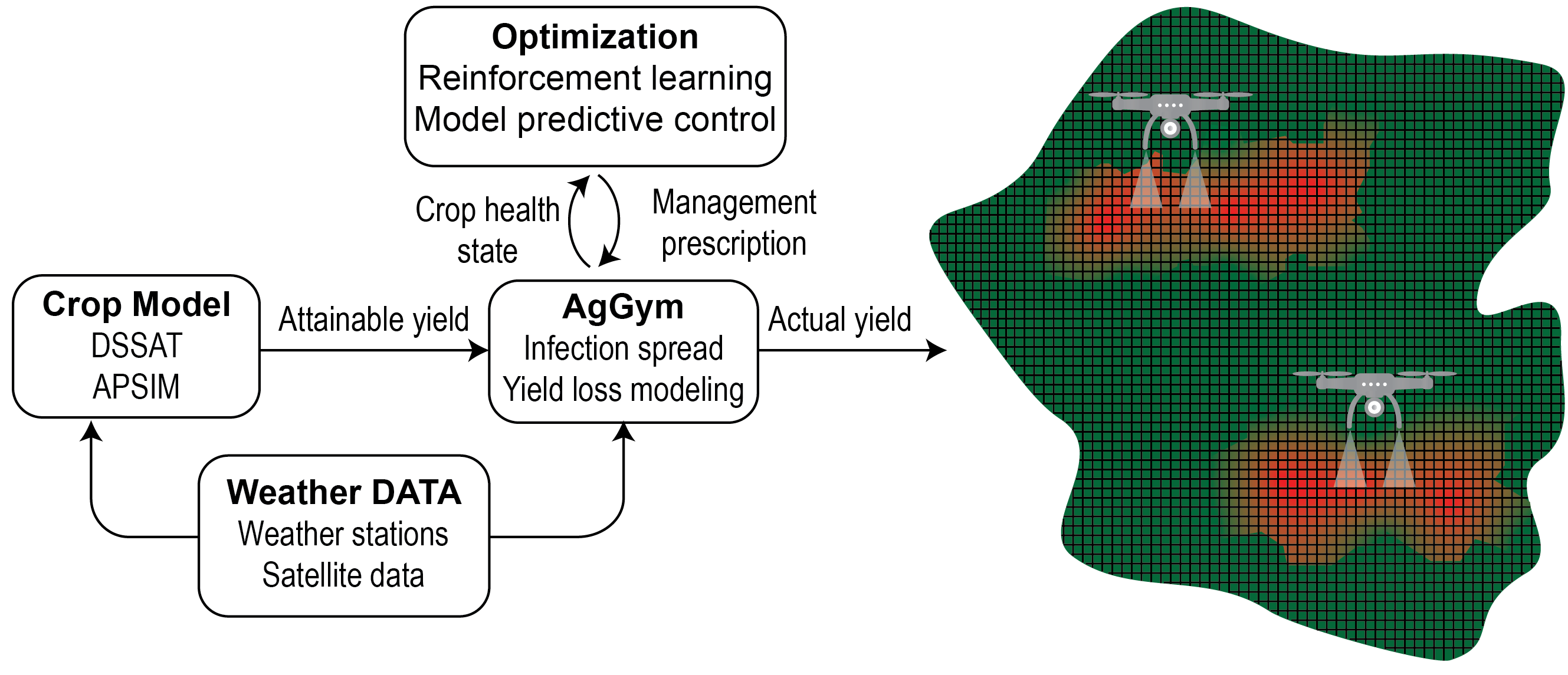}
	\caption{A high-level overview of interaction components with AgGym. The crop model and historical weather information inputs to AgGym provide a realistic field setup for initialization for threat dynamics simulation. After initialization, the deep RL agent trains via interacting with AgGym until a sufficient policy that maximizes the actual yield is achieved.}
	\label{fig:Flow}
\end{figure}

AgGym is a modular simulation framework that consists of two primary modules: (i) infection spread module, and (ii) yield estimation module under infection with and without local chemical treatment. As shown in Fig.~\ref{fig:Flow}, AgGym can be integrated with crop models and weather data. Crop models can provide AgGym with the attainable yield level based on factors such as weather conditions, soil nutrients and water availability. The AgGym module then attempts to estimate further yield loss due to biotic stresses. On the other hand, weather data that can be gathered from weather stations or satellites such as Earth Engine~\cite{gorelick2017google} and Microsoft Planetary Computer can be used by AgGym to estimate the severity of the biotic stresses that, in turn, determines the degree of yield loss. We envision that a primary use of AgGym is for designing optimal infection (pest/disease) mitigation strategies using algorithmic approaches that need to interact with the environment to obtain feedback on proposed infection control policies. In this paper, we used deep reinforcement learning (RL) approaches to design optimal mitigation strategies (e.g., scheduling in-season chemical treatments) which can be trained using the AgGym environment.

\subsection{Infection spread model}
\label{sec:infectionspread}
The plant science community has utilized models, such as Susceptible-Infected-Recovered (SIR) from human studies, to investigate the dynamics of plant disease spread ~\cite{arias2018epidemics}. The SIR model is a simple population-scale model that can track the number of disease susceptible individuals (units) in a population, the number of individuals infected with the disease, and the number of individuals who have recovered. The model assumes that individuals can only be in one of three states: susceptible (can become infected), infected (can recover), or recovered (immune to the disease). This model is not fully applicable to plant diseases; as models need to account for host (i.e., plant), pathogen or agent (i.e., fungi or bacteria), and environment, catering to the disease triangle \cite{horsfall1959plant}. In plant epidemiological studies, researchers use SIR models comprising differential equations for susceptible (S), infected (I), and recovered (R) populations, allowing investigations of simulated plant disease outbreaks in crops. An improved SEIR model defines categories for susceptible (S), latently infected (E), infectious (I), and post-infectious (R) individuals and have been utilized in fungal and viral diseases \cite{jeger2018plant,cunniffe2012time}. These mathematical models, though imperfect, are still helpful to plan mitigation strategies. 

In our proposed framework, we consider a ``simplistic" model that coincides with observed patterns in crop fields and prior disease spread modeling, where pathogen vectors are more likely to take the shortest path from plant to plant ~\cite{arias2018epidemics}. Taking into account the features of the disease and pest triangle, and the myriad of crops and diseases/pests that exist (viral, bacterial, fungal, insects) modeling plant disease or insect pests can become a daunting task \cite{britannica2024importance}. Focusing on disease and pest triangle - this model extends it to include IPM and reinfection probability after spraying. 
This further increases the issue of complexity by introducing chemical applications that differ in effectiveness, duration and pre-harvest-interval (PHI)~\cite{prodhan2018determination}. Due to the complexity of disease and insect-pest dynamics and their interactions with host and environment, we propose a modular approach. The modular aspect of this framework solves this problem by allowing  the community to alter and create models that simulate the spread of not only insect and vector transmitted disease (viruses) but any biotic disease of interest and predict the outcomes of IPM.

In AgGym, we divide a given field into multiple sub-regions and denote the health status for a sub-region $(i,j)$ by $H_{i,j}$. Therefore, the health status of the field can be represented as:

\begin{align}
    H_{l \times w} &=
    \begin{bmatrix}
    H_{1,1} & \cdots & H_{l,1} \\
    \vdots & \ddots & \vdots \\
    H_{1,w} & \cdots & H_{l,w}\\
    \end{bmatrix},
\end{align}
\\
where $l,w$ are the bounding lengths of the field. We can represent any arbitrarily shaped field into a set of rectangular areas. More details on how we handle arbitrary shaped fields can be found in the supplementary material (see Supplementary Material: Figure~\ref{fig:EndtoEnd}).. For simplicity, we discretize the health status into three levels, $\{h_1, h_2, h_3\}$ as defined below. 

\textbf{h1: Healthy} status signifies that the plants in a sub-region are not affected by a biotic stress or it has recovered from a prior stress.  

\textbf{h2: Infected} status signifies that the plants in a sub-regions are affected by biotic stress(es) and it also has the capability of spreading infection to its neighboring sub-regions.  

\textbf{h3: Degraded} status signifies that the sub-regions is maximally affected by biotic stress(es) with a loss of yield and is not recoverable by chemical treatments. For model development purposes, we assume that a degraded sub-regions no longer has the capability of spreading infection to its neighboring plants in the larger region. 

\begin{figure}
	\centering
	 \includegraphics[width=\linewidth,height=\textheight,keepaspectratio]{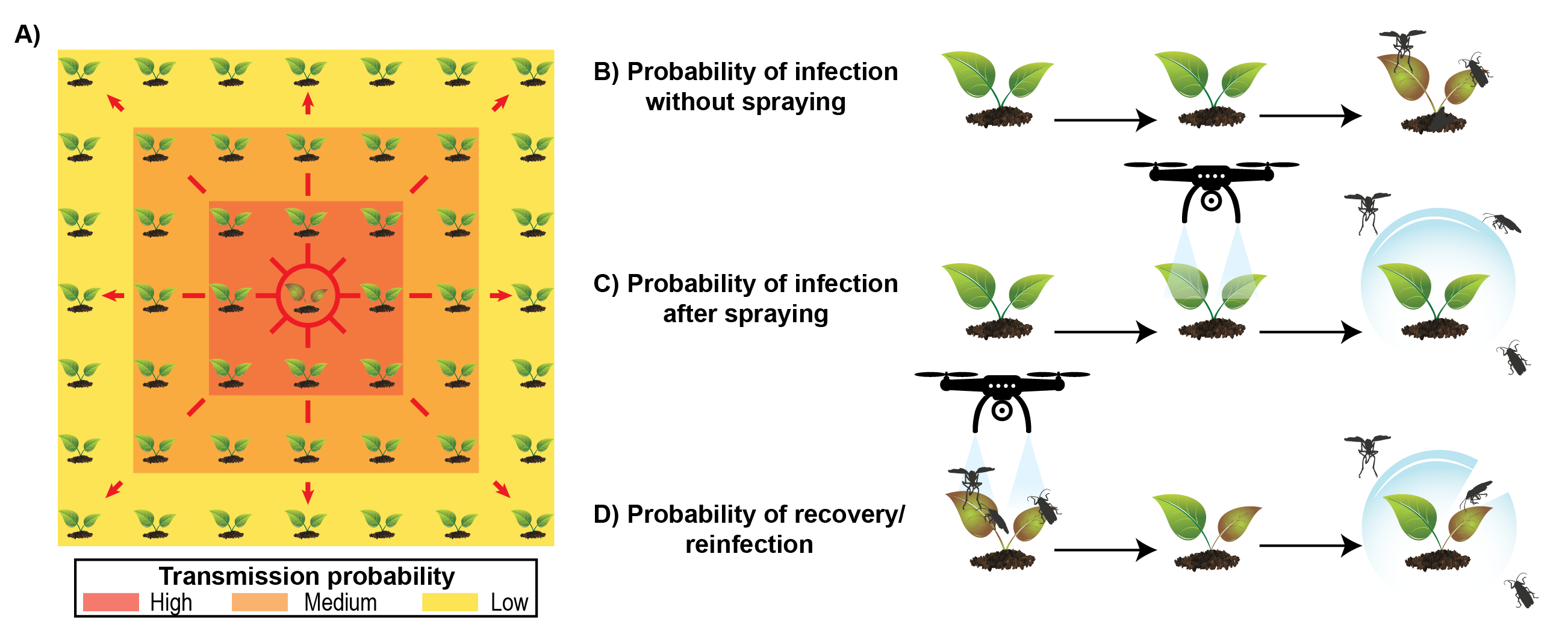}
	\caption{A) Three possible infection regions with High, Medium, and Low probability, indicate three levels of neighbors. Infection in these neighboring areas can occur under three conditions: 
    B) Probability of infection without prior spraying. C) Probability of infection with prior spraying. D) Probability of re-infection after prior spraying and recovery.}
	\label{fig:spread}
\end{figure}

Due to a modular framework, researchers can manipulate these scenarios as applicable to their host-disease or host-pest condition. 
With this setup, we define the probability of infection spread, denoted by $PIS(P_{i,j}|P_{k,l})$, as the probability of infection initiation at a sub-region $P_{i,j}$ due to an infected sub-region $P_{k,l}$. The probability of infection spread is a function ($S$) of the physical distance between the two sub-regions, $r$ and the environmental factors, $E$. 
\begin{equation}
    PIS(P_{i,j}|P_{k,l}) = S(r,E)
\end{equation}
Depending on the biotic stress characteristics, there can be various environmental factors that may affect the infection spread between neighboring plants in a larger region such as terrain, soil distribution, and wind flow characteristics. For example, studies have identified Stripe Rust \textit{Puccinia striiformis f. sp. tritici} epidemic infections based on wind patterns in wheat~\cite{chen2005epidemiology}. However, there is still a lack of biophysical understanding and mathematical formulations of how these environmental factors may impact local spread of biotic stresses. Hence, in this early attempt of building a biotic stress simulator, we ignore the effect of environment in local stress propagation. Similarly, the dependency of infection spread on physical distance may also manifest in various ways. In the current implementation, we consider a simple and intuitive model where, $PIS$ becomes lower with increase of distance from the infected plants in a sub-region. Based on this assumption, we define three neighborhood zones with increasing distance from the infected sub-regions, with three discrete possible values of $PIS$, $\{s_H,s_M,s_L\}$, where the subscripts signify $H = High, M = Medium, L = Low$ (see Fig.~\ref{fig:spread} (A)). Both the neighborhood sizes and $\{s_H,s_M,s_L\}$ values are user-defined parameters. Further, we note that given our modular software architecture, the user can also import other $S(r,E)$ functions in the framework that could be more suitable for specific use cases.  

Upon defining $PIS$, we now introduce the probability of infection, $PI(P_{i,j})$ for a given sub-regions $P_{i,j}$. Other than depending on the infection transmission probability from neighbors, $PI(P_{i,j})$ may also depend on the history of previous infections, and the history of chemical treatments. Both prior infections and chemical treatments may reduce the susceptibility of future infections. To capture these effects, we consider an additional scaling factor, $0 < \lambda \leq 1$ such that
\begin{equation}
    PI(P_{i,j}) = \frac{\lambda}{|\{Nb(P_{i,j})\}|}\sum_{P_{k,l}\in \{Nb(P_{i,j})\}}PIS(P_{i,j}|P_{k,l})
\end{equation}
where, $\{Nb(P_{i,j})\}$ is the set of neighboring plants in a region of $P_{i,j}$ as defined by the user. In the current implementation, we consider (1) $\lambda = \lambda_{RI}$ ($RI$ stands for reinfection), when sub-region $P_{i,j}$ had a history of previous infection that it recovered from (possibly due to chemical treatment); and (2) $\lambda = \lambda_{Spray}$, when sub-regions $P_{i,j}$ had been treated previously without any infection. While these are user-defined parameters, typically, we choose $\lambda_{RI} < \lambda_{Spray} < 1$. In all other cases, we consider $\lambda = 1$. All these different possibilities are illustrated in Fig.~\ref{fig:spread} (B,C,D).




\subsection{Yield loss and management}
From a crop production viewpoint, farmers are interested in producing a high-yielding, high-quality crop with optimal resource utilization. For practical application of IPM strategies, researchers can look at the issue of pathogen-induced crop losses from the perspective of potential yield $>$ attainable yield $>$ actual yield (farmer's realized yield). Yield-defining factors, such as sunlight, temperature, rainfall, crop phenology, and physiological status, are crucial for determining potential yield, while yield-limiting factors, such as water and nutrients, are vital for attainable yield. Pathogens, pests, and weeds are considered yield-reducing factors. In developing pest/pathogen development and spread dynamics of yield-reducing factors, we include factors that govern potential and attainable yield. We also established plant health status that vary with health grades, from healthy to infected to degraded, with varying probabilities of infecting neighbors and getting re-infected. It ensured that the model accounts for the foundations of the disease and pest triangle axes.

AgGym also includes the interaction of the type of pathogen, current plant phenology stage at initial infection, and the duration of infection~\cite{madden2000coupling}. This builds the framework to accurately model the yield loss by taking into account the type of damage (e.g., target plant organ), how much damage (e.g., duration), and when it occurs (e.g., phenology stage)~\cite{cooke2008ecological,ficke2018understanding}. To illustrate, an infection that targets reproductive organs during flowering will cause more yield loss than an infection that targets leaf tissue during grain filling stage. 

\begin{figure}
	\centering
	 \includegraphics[width=\linewidth,height=\textheight,keepaspectratio]{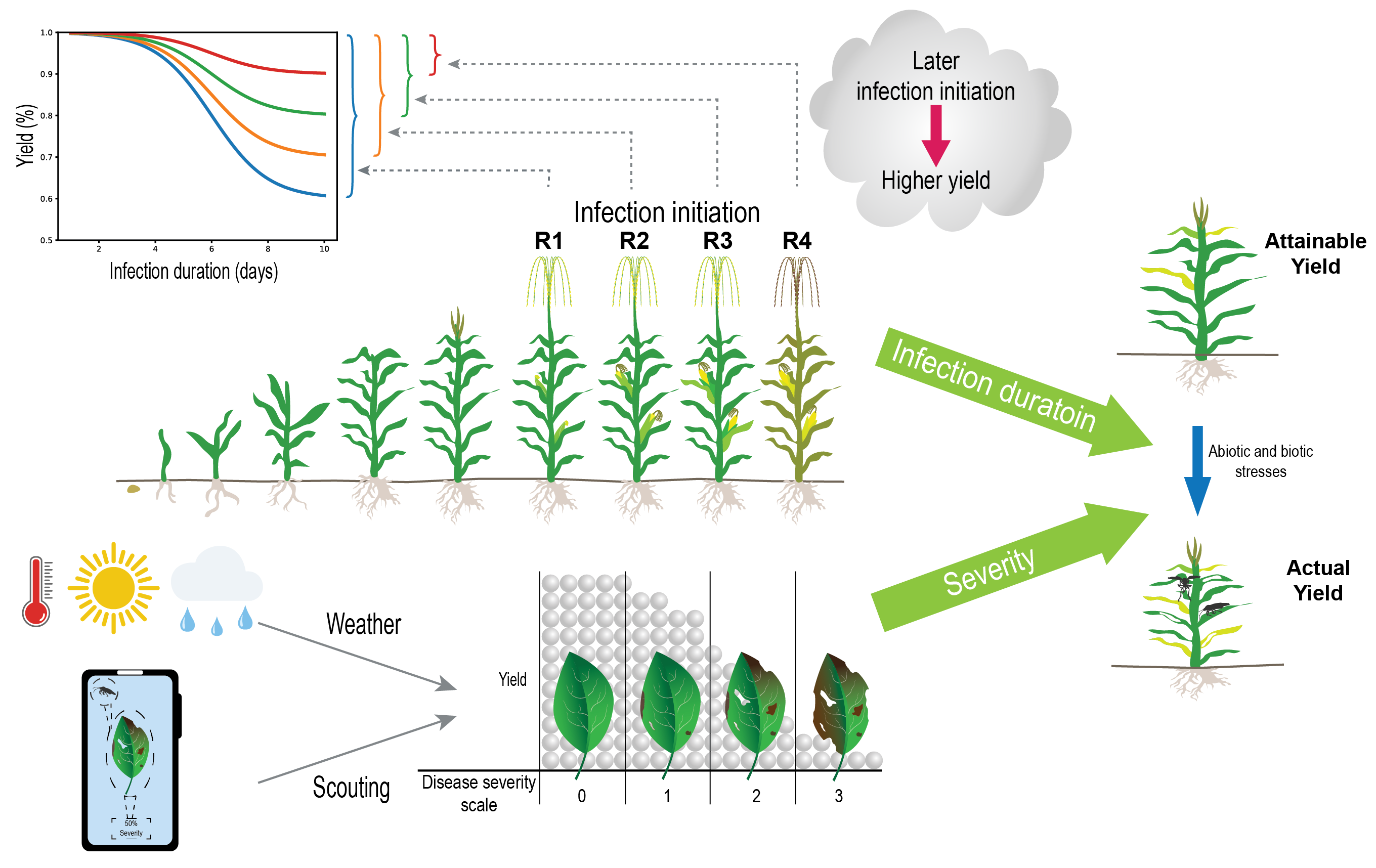}
	\caption{Yield loss as a function of infection severity and time elapsed since the initial day of infection. Another factor influencing yield loss is the specific time point during the growing season (R1-R7) when the infection occurs.}
 \label{fig:Yloss}
\end{figure}

Formally, let the actual yield ($Y_{act}$) be a fraction ($\eta_{Y}$) of the attainable yield ($Y_{att}$) due to the various stresses. We consider that the yield reduction is a function of duration of infection ($T_{inf}$) and the growth stage at the time of infection initiation ($G_{inf}$).
As shown in Fig.~\ref{fig:Yloss} and suggested `damage curve' in literature~\cite[p.~306]{singh2010soybean}, we choose an inverse sigmoid function form for $g(\cdot)$ that can capture how the actual yield may reduce as a function of the duration of infection as a fraction of the attainable yield. As the infection initiation occurs at a later growth stage, the function shifts upwards signifying a lower yield loss. Therefore,
\begin{equation}
    \eta_Y = Y_{min}(G_{inf}) + (1 - Y_{min}(G_{inf})) \frac{exp^{-T_{inf} + T_{50\%}}}{1 + exp^{-T_{inf} + T_{50\%}}}
\end{equation}
where, $Y_{min}(G_{inf})$ is the minimum actual yield obtained without any treatment for the stress which is a function of $G_{inf}$. $T_{50\%}$ signifies the infection duration by which there will be $50\%$ of the total yield loss. This function represents the yield decay of a crop after being infected and left untreated for a given $G_{inf}$. 

The yield loss ($Y_{loss}$) can be further impacted by the severity of infection. Stress severity can be a function of the weather parameters. For example, historical studies of \textit{Sclerotinia sclerotiorum} severity in soybean show that severity can be represented as a function of precipitation and average air temperature ~\cite{fall2018case}. Severity can also be quantified via scouting of stresses by experts or other means. In AgGym, we provide the flexibility of choosing either of these approaches to include the severity factor in simulation as shown in Fig.~\ref{fig:Yloss}. 
The dependency of yield loss on severity can be determined based on domain knowledge about specific stresses for the crop of interest. In the current implementation, we assume a simple multiplicative decomposition as follows:
\begin{equation}
   Y_{loss}  = (1 - \eta_Y) \times \mathcal{S}_{inf}
   \label{eq:yield_loss}
\end{equation}
where, $\mathcal{S}_{inf} = [0,1]$ is the (normalized) severity index. The yield loss is maximum for the highest severity index value $1$ and there is no yield loss for the lowest severity index value $0$.

In this study, we deliberately incorporated varying levels of pesticide efficacy to simulate real-world conditions. By including different efficacy levels, we aimed to replicate the diverse range of pesticide effectiveness that is typically observed in agricultural practices. This approach was adopted to ensure that our findings are more representative of actual field scenarios, where pesticide treatments may exhibit a spectrum of efficacy. By incorporating a mix of highly effective, moderately effective, and less effective pesticides, we aimed to capture the complexities and challenges faced by farmers when dealing with pest management in real-world agricultural settings. This study provides a more comprehensive understanding of the potential outcomes and limitations of different pesticide efficacy levels, allowing for more informed decision-making and practical recommendations for pest control strategies in agricultural systems.

Formally, we denote the probability of infection recovery of an Infected sub-region (\textbf{h2}) after application of chemical treatment, by $PIR$. As discussed above, we consider three levels of effectiveness for chemical treatments, namely, highly effective (HE), moderately effective (ME), and less effective (LE). While $PIR$ values can be chosen via calibration using real data of specific scenarios, we assume that $PIR$ decreases monotonically as effectiveness of chemical treatments goes down. The cost of chemical treatments $C$ may also vary depending on the effectiveness levels and we keep that provision in the AgGym implementation. 

All these parameters described above can be adjusted by the users of AgGym. 
\FloatBarrier
\subsection{Supervisory decision-making}\label{RL}
AgGym framework is multi-faceted, and one of the uses includes the designing of optimal and precise farm management strategies. Here, we present an example use case where the AgGym simulation platform is used to develop an optimal schedule for chemical treatment in a field during an entire growing season. The goal here is to provide decision support to a farmer on `when to spray' (i.e., spraying schedule) and `what to spray' (i.e., type of chemical), given a perfect observation of the infection status of the field. We also assume that the infected plants in a sub-regions are sprayed accurately based on the spraying schedule using the suggested type of chemical. We acknowledge that some of these assumptions, such as perfect knowledge of infection and perfect actuation of spraying can be restrictive. However, this initial proof-of-concept study still demonstrates the potential of a supervisory decision-making framework developed using AgGym. Future work by the broader research community can build on this work to consider more realistic constraints and relax some of the assumptions made here. 

Specifically, we formulate a model-free deep reinforcement learning (RL) problem to design the supervisory decision-making framework described above. Reinforcement learning (RL) or deep reinforcement learning (DRL, RL that leverages deep neural networks to handle complex environment and action structures) aims to produce optimal sequence of actions (decisions) in an environment to obtain maximum expected reward as defined by the user. In our context, we aim to obtain optimal spraying schedule (action), given the infection status of a field (environment), to obtain maximum possible yield recovery with minimum amount or cost of chemical (reward). Mathematically, RL (or DRL) is formulated by a Markov Decision Process (MDP) described by a 4-tuple $(S, A, T, R)$, where:

\begin{itemize}
    \item $S$ represents the set of possible states of the environment. 
    \item $A$ represents the set of all actions available to the RL agent.
    \item $T: S \times A \times S' \rightarrow [0, 1]$ is the transition function, indicating the environment's probability of transitioning from one state $S$ to another state $S'$ due to an action $A$.
    \item $R: S \times A \times S' \rightarrow R$ is the reward received by the agent for a transition from state $S$ to state $S'$ under the specific action $A$.
\end{itemize}

The objective of RL (or DRL) is to design an optimal policy $\pi^*$ that enables the agent to maximize the expected reward along a trajectory $\tau$ which is a sequence of states and actions $(s_0, a_0, s_1, a_1, ..., s_T, a_T)$. The expected reward is calculated as $J(\theta) = \mathbb{E}_{\tau \sim \pi_\theta}[\sum_{t=0}^{T} \gamma^t R(s_t, a_t, s_{t+1})]$, where $\theta$ signifies the parameters of the policy, and $\gamma$ the discount factor. In the case of a DRL, the policy is modeled by a deep neural network. Hence, $\theta$ denotes the parameters of the deep neural network that are learned by interacting with the environment. In addition, we also define `horizon' as the duration over which the performance of a policy is assessed, and an `episode' that describes a complete trajectory. With this setup, we describe the state, action and reward that we consider for our problem. 

\textbf{RL for optimal spray scheduling}
\label{sec:rl_spray_scheduling}

In this paper, we consider the health status matrix $H$ of a field as the state observation. Other auxiliary information about the environment, such as weather parameters over the growing season, can be also be used as a part of the state observation, depending on user customization. In our implementation, we also allow the user to only consider certain manageable areas in the field (as opposed to the whole field) that may signify specific regions of the field with a historically high risk of infections.

We consider the set of possible actions as $a_{\pi} = \{NO, LE, ME, HE\}$ denoting the `no spray' ($NO$) decision, followed by spraying chemicals with different levels of effectiveness (and possibly with different cost), as described earlier. While we consider this specific discrete action formulation in this study, users can customize the action space with a simpler (e.g., binary action - spray or no spray) or a more complex (e.g., continuous action - how much to spray) depending of the demand of a specific application. 

The objective of the RL agent is to mitigate the impact of biotic stresses and increase revenue. Hence, components of the reward functions are tightly related to the net revenue generated from each unit. Therefore, we consider an objective function that minimizes both yield loss as well as cost of chemical treatment. Specifically, we define the reward function, $r_{1}$, for the yield loss minimization as:
\begin{equation}
    r_Y = - Y_{loss} \times UAY \times PPB
\end{equation}
where, $UAY$ is the unit attainable yield and $PPB$ is the price per bushel of yield. On the other hand, the reward function, $r_{2}$, for the cost of chemical minimization is defined as:
\begin{equation}
    r_C = - |h_2(H)| \times C(a_{\pi}) \times UPP
\end{equation}
where, $|h_2(H)|$ is the infection count of the field, $C(a_{\pi})$ is the cost of chemical chosen by the policy $\pi$, and $UPP$ is the unit pesticide price. Note that $r_C$ is zero when either there is no infection in the field or the RL agent suggests `no spray'. Also, we reiterate that we assume a perfect observation of health status of the field, and spray is applied to only the infected sub-regions when the RL agent decides to spray.  

The overall reward function $r$ is expressed as $r = r_Y + r_C$ with the goal achieving $r$ as close to zero as can be attained based on a given infection rate and severity. Similarly, with the State-Action formulation, the reward formulation is also customizable in a modular fashion. 
\section{Results and Discussion}
We have performed extensive validation of the different modules of the proposed AgGym platform to demonstrate that it has the capability to emulate real agricultural systems with careful calibration. We also present a simulation-based example use case to show that the RL-based management may be able to reduce the cost and extent of chemical treatment while maintaining or improving yield recovery.  
\subsection{Stress simulation validation}
We begin with the validation of the infection spread model that is critical for AgGym's 
applicability in real-world agricultural scenarios. In this section, we consider the spread of an insect pest in an agricultural field. We focus on evaluating the capability of AgGym to accurately capture both the spread of insects as well as the resulting impact on crop yield. 
This validation enables us to devise effective mitigation measures and facilitate evidence-based decision-making for implementing successful pest management strategies using AgGym.

\textbf{Validation of insect pest spread} 

\begin{figure*}[t!]
\centering
\includegraphics[width=\linewidth,height=0.9\textheight,keepaspectratio]{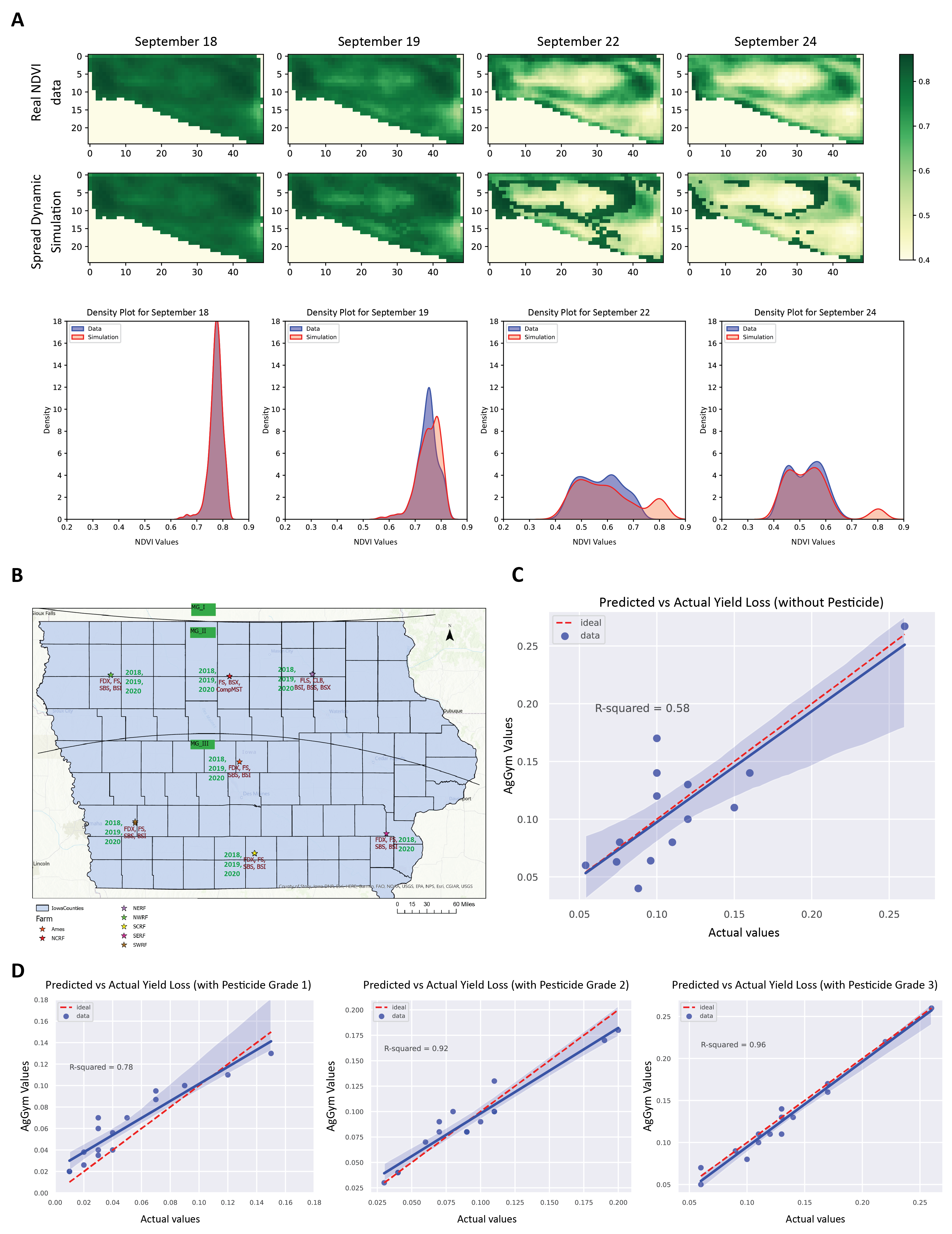}
\caption{A) Comparison of infection spread during the growth season using Normalized Difference
Vegetation Index (NDVI) between simulation data and real data.(Comparative Visualization of Disease Spread Dynamics during the growth season through NDVI and Distribution Functions.) B) Disease and severity indices information related to different agricultural fields in Iowa counties from 2018 to 2020. The MG I, II, III along with the curved lines show the regions in Iowa where these maturity groups are grown. C) R-squared values for yield loss predictions without pesticide applications and D) with different Grades of pesticide application.}\label{fig:infection}
\end{figure*}

We present the validation process of our spread dynamic model using satellite imagery and the Normalized Difference Vegetation Index (NDVI) obtained from the satellite data provider, Planet~\cite{planet}. The main objective of this validation study is to evaluate the simulation model's capability to accurately replicate the observed spread patterns in real-world agricultural fields. We consider satellite data from an alfalfa field located in East Central Iowa that encountered an infestation of fall armyworm (\textit{Spodoptera frugiperda}) the fall of 2021. Fall armyworm is an invasive pest of global importance that feeds on green leaves and other parts of a plant, causing severe agricultural losses~\cite{overton2021global}.
We analyzed the NDVI values for a specific time interval during the growing season. Typically, the NDVI value decreases as the fall armyworm damages the crop and hence, we considered areas with NDVI values below a threshold ($0.7$ in this paper) as infested regions. 
By incorporating these initial infested sections, our simulation model was able to simulate the spatial spread of infestation over time. The simulated NDVI values were then visualized alongside the real NDVI values from satellite data for the corresponding time intervals. Through a comparison of the NDVI trends from both sources, we can directly assess the accuracy and reliability of our spread dynamic model in capturing the spread of insect pests during the growing season.
Figure~\ref{fig:infection} provides a graphical representation of the comparison between our simulation data and the real data (Figure~\ref{fig:infection} (A)). The spatial distributions of predicted NDVI values from our simulation model are compared with those obtained from the actual satellite data. Visual inspection suggests the close alignment between the sets of plants in two sub-regions. While the actual data is smoother compared to the simulated data, the overall trends affirm that the AgGym infection spread module successfully emulates the observed patterns of pest propagation in the field for this case study. Alongside the spatial distribution, Figure~\ref{fig:infection} (A) also includes density plots of the observed (blue) and simulated (red) NDVI values, which further delineate the distribution functions of these datasets. In some sense, these distributions quantify the severity and extent of the infestation. Despite slight visual differences, the plots show that the central tendency and variability between observed and simulated data are closely aligned, underscoring the model’s effectiveness in capturing the dynamics of disease spread.
\FloatBarrier

\textbf{Validation of yield loss and mitigation}

We conducted an extensive analysis using real-world agronomic data collected from seven distinct farms in Iowa: Ames, NCRF (Kanawha), NERF (Nashua), NWRF (Sutherland), SCRF (McNay), SERF (Crawfordsville), and SWRF (Armstrong) between the years 2018-2020. The dataset from each farm comprised information on soybean yields and various stress/disease indices recorded, as represented in Fig~\ref{fig:infection}-(B). The description of various crop stress and disease indices are provided in the supplementary material (see Supplementary Material: \hyperref[sec:agronomic_data_description]{Agronomic Data Description}). This information was collected under conditions with no pesticide application, and following the implementation of 19 various treatment regimens. These treatments included different fungicide products, their Fungicide Resistance Action Committee (FRAC) groups (indicating their mode of action), application rates, and the companies that manufacture them. Examples of the treatments include Miravis Neo (FRAC: 3,7,11; 13.7 oz/A; Syngenta), Domark 230 (FRAC: 3; 5 oz/A; Gowan), and Quilt Xcel (FRAC: 3,11; 10.5 oz/A; Syngenta) as shown in Table~\ref{tab:comparison}.
The attainable soybean yield for each farm in each respective year was determined by taking an average of the yields linked to the top six treatments. From the attainable yield and the actual yield data, we were able to compute the yield loss percentages under each scenario.

Our simulation model's yield loss prediction relied on the severity index of each field. As such, we derived the severity index for each farm by calculating the mean of all the severity indices noted for that particular farm.
Furthermore, in our quest to validate our model's ability to account for different efficacy grades of pesticide application, we needed to compare it with real-world yield loss data obtained from treatments of varying effectiveness. Therefore, we categorized treatments based on their efficacy levels and computed the corresponding yield loss and severity indices.
\FloatBarrier
We subsequently ran our simulation model for all farms through the years, sans pesticide application, and juxtaposed the results with the real-world data. The computed R-square value of 0.58 (Fig~\ref{fig:infection}-(C)), indicating a significant fit, underscored the model's effectiveness.
To better mimic real-world conditions, we applied different grades of pesticide, ranging from intensive to minimal, exclusively at stage R3 (Reproductive Stage; beginning pod) across the entire field. This stage was consistently represented in all the real data sets under consideration. The treatments were categorized into three efficacy grades based on yield loss and initial disease severity prior to application. Grade 1 represented the highest efficacy, characterized by minimal yield loss and effective control in sub-regions with severe initial disease conditions. The efficacy level decreases gradually for Grades 2 and 3. This categorization allowed for precise validation of our simulation model's ability to predict outcomes based on varying treatment intensities under diverse pre-treatment disease pressures. Our model's predictive capability is demonstrated by high R-square values of 0.78, 0.92, and 0.96 corresponding to pesticide grades 1, 2, and 3 respectively, as depicted in Fig~\ref{fig:infection}-(D). A comparison of the predicted yield loss across these scenarios with the actual field data allowed us to evaluate the accuracy and reliability of our simulation model in forecasting yield losses under an array of pest management strategies. The resulting insights into the efficacy of different pesticide application techniques were invaluable. They underscored the potential benefits of leveraging our simulation model to optimize pest management decisions and mitigate yield losses across agricultural fields.
\begin{figure*}[t!]
\centering
\includegraphics[width=\linewidth,height=\textheight,keepaspectratio]{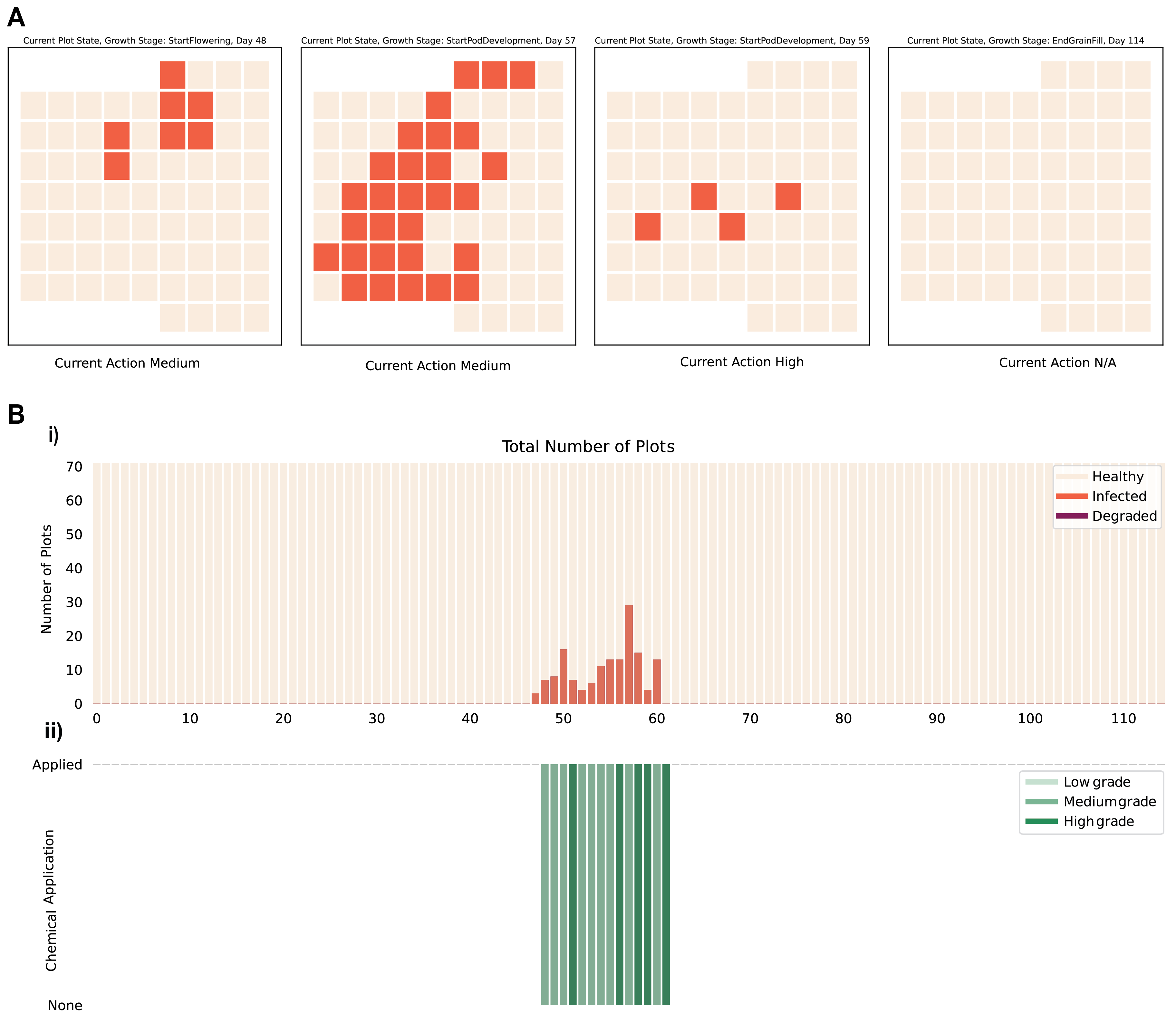}
\caption{(A) Spread dynamics of infection during different growth stages. (B-i) Number of infected plots and (B-ii) pesticide applications on the simulation day, shown separately. The darker a section, the more efficient and expensive the pesticide application.}\label{fig:side}
\end{figure*}
\subsection{Precision management planning}
Upon validating the AgGym simulation modules, we demonstrate its use for designing optimal and precise farm management strategies. Specifically, we present a few benchmark model free deep RL policies trained with AgGym and evaluate their effectiveness in the context of our real life use cases. 

\textbf{Deep RL policy learning with AgGym}

To demonstrate AgGym's utility as a RL gym environment, We deployed three deep RL strategies: the Double Deep Q Network (DDQN) \cite {van2016deep}, a Q-learning-based method, and two prominent policy gradient algorithms, Proximal Policy Optimization (PPO) and Trust Region Policy Optimization (TRPO) \cite{schulman2015trust}, all implemented using Preferred Network’s Deep RL library, PFRL \cite{JMLR:v22:20-376}. In the experimental configuration, each episode was set to span 115 days, representing a typical crop growing season (for row crops in the Midwest of United States). Within this setup, the initiation of infection was programmed to occur during the reproductive stage, typically between time steps 44 to 55. Our evaluation and comparison of these algorithms were based on the average reward obtained during training over a window of 10,000 episodes. We conducted ten training runs with different random seeds for the growing season setting. More details of the experimental parameters can be found in the SI.

Insights into the algorithms' performance can be derived from infection and spraying action sub-region over the course of the growing season (i.e, one episode for RL agent). As an example, Fig.~\ref{fig:side} showcases the performance evaluation of the DDQN algorithm, based on the most optimal seed selected from ten training iterations, conducted on a farm near Ames, Iowa. In (A), the spatial dynamics of an infection is depicted, illustrating how infections emerge, spread, and are subsequently managed. Infected sub-regions are shown being sprayed and returning to health across various stages of the growing season. Additionally, (B-i) and (B-ii) detail the number of infected sub-regions and the different grades of pesticide application on these sub-regions, respectively, with darker sections indicating more efficient but costlier pesticide usage. A significant observation was the absence of any degraded or infected plants in a sub-region in the final phase preceding the harvest time termination. This clearly demonstrates the agent's capability to effectively apply pesticides in response to the detected threats, showcasing the system's responsiveness and precision in managing biotic stress. Note that we assume perfect observation of the infection state of the field and perfect actuation, i.e., error-free spraying of chemicals for these experiments. Results for different RL algorithms for different runs are summarized in the SI. Notably, the agent trained using TRPO showed superior performance compared to the others, achieving the maximum expected reward with the lowest standard deviation.

\textbf{Schedule-based vs. RL-based management}

We conducted a comparative analysis to evaluate the effectiveness of RL-based management strategies against the conventional agricultural practices in pest management. As the TRPO algorithm demonstrated the most promising results in terms of cost-effectiveness, we used the TRPO-generated policies for this analysis. The real-world scenarios employed for this evaluation (that were previously utilized for validating yield loss and mitigation strategies) encompass various fields such as Ames, NCRF (Kanawha), NERF (Nashua), NWRF (Sutherland), SERF (Crawfordsville) for the year 2018, SWRF (Armstrong) for 2019, and SCRF (McNay) for 2020. The objective is on contrasting the conventional approach of uniform chemical application across the entire field with the precision strategy employed in the RL setting, where chemical is applied only as necessary to infected sub-regions. This approach is quantified in the \% Sprayed column of Table~\ref{tab:result}, showing the percentage of the field that received pesticide applications in the RL scenario, in contrast to the full-field application in real-world practices.

The results, as detailed in Table~\ref{tab:result}, show the potential gain for using the RL-based approach. Conventional practices typically involve extensive chemical application, leading to significant costs and yield losses. For instance, the traditional approach in the 2018 Ames scenario resulted in a pesticide cost of \$2.14 per acre and an 18\% yield loss. Conversely, the RL method, employing the TRPO algorithm, achieved a significant reduction in costs to \$0.17 per acre and completely eliminated yield loss, with pesticide applications required on only 4.22\% of the field. Such substantial decreases in chemical usage and yield loss across all fields investigated underscore not only the economic benefits of the optimization (RL in this paper) approach but also its potential to significantly enhance the sustainability of agricultural management.

\begin{table*}
\centering
\caption{Comparison of Real data and AgGym (RL) result}
\small
\resizebox{\textwidth}{!}{%
\begin{tabular}{lcccccccc}
\toprule
\textbf{Location} & \multicolumn{4}{c}{\textbf{Real data, typical management (100\% Sprayed)}} & \multicolumn{4}{c}{\textbf{AgGym, RL management}} \\
\cmidrule(lr){2-5} \cmidrule(lr){6-9}
 & \begin{tabular}[c]{@{}c@{}}Pesticide\\ cost (\$/Acre)\end{tabular} & \begin{tabular}[c]{@{}c@{}}Yield\\ loss (\%)\end{tabular} & \begin{tabular}[c]{@{}c@{}}Yield\\ cost (\$)\end{tabular} & \begin{tabular}[c]{@{}c@{}}\%\\ Sprayed\end{tabular} & \begin{tabular}[c]{@{}c@{}}Pesticide\\ cost (\$/Acre)\end{tabular} & \begin{tabular}[c]{@{}c@{}}Yield\\ loss (\%)\end{tabular} & \begin{tabular}[c]{@{}c@{}} Yield\\ cost (\$) \end{tabular} & \begin{tabular}[c]{@{}c@{}}\%\\ Sprayed\end{tabular} \\
\midrule
Ames (Ames)-2018 & 2.14 & 18 & 26.3 & 100 & 0.17 & 0 & 0 & 4.22 \\
Armstrong (SWRF)-2019 & 2.44 & 5 & 9.76 & 100 & 0.17 & 0 & 0 & 3.7 \\
Crawfordsville (SERF)-2018 & 2.29 & 7 & 11.64 & 100 & 0.34 & 0 & 0 & 7.9 \\
Kanawha (NCRF)-2018 & 1.86 & 11 & 11.84 & 100 & 0.23 & 0 & 0 & 6.45 \\
Nashua (NERF)-2018 & 2.23 & 8.4 & 12 & 100 & 0.62 & 0 & 0 & 14.86 \\
Sutherland (NWRF)-2018 & 2.43 & 5 & 10 & 100 & 0.23 & 0 & 0 & 4.94 \\
McNay (SCRF)-2020 & 1.23 & 11 & 8.12 & 100 & 0.17 & 0 & 0 & 7.32 \\
\bottomrule
\end{tabular}}
\label{tab:result}
\end{table*}
\FloatBarrier
\section{Conclusion}
The new paradigm of cyber-agricultural systems (CAS)~\cite{Sarkar2024-CAS} is bringing advances in sensing, modeling, and actuation to agriculture to enable ultra-precision chemical application that can lead to increased yields, reduced resource utilization, and enhanced environmental sustainability. In this paper, we presented an open source and modular simulation platform for modeling and analysis of biotic stresses in row crop agriculture at the field-level. We envision that a community of users will be able to leverage the AgGym platform for simulating `what-if scenarios' after calibrating for specific field geometries, weathers, crops, and stress types. Specifically, we note that the modular nature of our framework allows for modifying (e.g., changing the function structures, adding/removing variables) different parts of the simulator as needed. An important use of AgGym would be for designing optimized strategies for mitigating the impact of stresses. While agronomists, agricultural extension specialists, and farmers may find this useful, we anticipate that AgGym will also serve as a benchmark use case for the optimization and machine learning research community.

As indicated earlier, AgGym is an early attempt towards simulating biotic stresses and their impacts at a field-scale. Hence, there is a significant scope of improvement such as including impacts of additional environmental factors (e.g., wind, terrain) and historical data (useful for certain stresses that have strong year-to-year correlation). In addition, there is a wide range of variability of how infection spreads and affects crop yield depending on crop and stress types. Hence, it is very likely that more complex modeling could be involved to build a high-fidelity simulator for specific use cases. Future directions for the supervisory decision-making part will include consideration of sensing imperfections, actuation uncertainties while planning for optimal spraying under resource constraints.  

\section*{Acknowledgments}
This work was supported by the COALESCE: COntext Aware LEarning for Sustainable CybEr-Agricultural Systems (CPS Frontier \#1954556), AI Institute for Resilient Agriculture (USDA-NIFA \#2021-647021-35329), Smart Integrated Farm Network for Rural Agricultural Communities (SIRAC) (NSF S \& CC \#1952045), Iowa Soybean Association, USDA CRIS project IOW04714, RF Baker Center for Plant Breeding, and Plant Sciences Institute. We also thank the members of the Iowa State University Soynomics team for their valuable contributions in research data generation, particularly Stith Wiggs and Yuba Kandel for their data collection efforts at the research farms. Finally, we thank a broader community of agronomists, plant pathologists and entomologists who provided critical feedback on this work.






\bibliographystyle{unsrtnat}
\bibliography{references}  
\newpage
\thispagestyle{empty}

\begin{center}
    \rule{\linewidth}{0.5mm} \\ 
    \vspace{0.4cm}
    \LARGE\textbf{Supplementary Information} 
    \vspace{0.4cm} \\
    \rule{\linewidth}{0.5mm} 
\end{center}
\renewcommand{\thesubsection}{\Alph{subsection}}
\setcounter{subsection}{0} 
\subsection{Materials}
\subsubsection{Agronomic data description}
\label{sec:agronomic_data_description}

This section provides detailed descriptions of the various soybean stress and disease indices used in our validation. These indices are critical for understanding the health and productivity of crops (soybean in this case) in different agronomic conditions.

\textbf{Foliar Disease Index (FDX)}:
The Foliar Disease Index (FDX) is an index for foliar sudden death syndrome (SDS) in soybean, based on the disease incidence (DI) and disease severity (DS). It is calculated as FDX = (DI * DS) / 9 \cite{Gibson1994-ua}.

\textbf{Frogeye Leaf Spot (FLS)}:
Frogeye Leaf Spot (FLS) is a foliar fungal disease caused by Cercospora sojina that primarily affects the foliage of soybean, although lesions can appear on pods, stems, and seeds \cite{Mian2008-ck}.

\textbf{Septoria Brown Spot (SBS)}:
Septoria Brown Spot (SBS) is a fungal disease caused by Septoria glycines, which develops on leaves and can cause severe canopy defoliation\cite{Lim1980-kl}.

\textbf{Septoria Brown Spot Severity (BSS)}:
Septoria Brown Spot Severity (BSS) is a visual rating of the proportion of symptoms present in the canopy, with higher values indicating a greater amount of affected canopy\cite{Hartman2015-pc}.

\textbf{Septoria Brown Spot Incidence (BSI)}:
Septoria Brown Spot Incidence (BSI) is the rating of the amount and location of septoria brown spot symptoms in the canopy.

\textbf{Septoria Brown Spot Index (BSX)}:
The Septoria Brown Spot Index (BSX) is based on the disease severity (BSS) and disease incidence (BSI).
\subsubsection{Comprehensive Overview of Fungicide Treatments}
Comprehensive details on the fungicide treatments, including product names, FRAC groups, companies, application rates, and timing, are presented in the Table~\ref{tab:comparison}. These tables complement the detailed information about the various treatment regimens discussed in the main text.

\FloatBarrier
\begin{table*}[h]
\centering
\caption{Comparison of Treatments Used in Different Years}
\label{tab:comparison}

\begin{center}
    \textbf{Fungicide Trial in Year 2018}
\end{center}
\begin{adjustbox}{max width=\textwidth}
\begin{tabular}{cccccc}
\toprule
\textbf{Trt} & \textbf{Product} & \textbf{Company} & \textbf{Rate} & \textbf{Timing} \\
\midrule
1 & UTC & --- & --- & --- \\
2 & Miravis Neo & Syngenta & 13.7 oz/A & R3 \\
3 & Quilt Xcel & Syngenta & 10.5 oz/A & R3 \\
4 & Topguard EQ & FMC & 5 oz/A & R3 \\
5 & Restricted & Restricted & Restricted & Restricted \\
6 & Restricted & Restricted & Restricted & Restricted \\
7 & Lucento & FMC & 5 oz/A & R3 \\
8 & Aproach Prima & DowDuPont & 6.8 oz/A & R3 \\
9 & Aproach & DowDuPont & 6 oz/A & R3 \\
10 & Delaro + Induce & Bayer & 8 oz/A & R3 \\
11 & Fortix & Yuba & 5 oz/A & R3 \\
12 & Priaxor & Yuba & 4 oz/A & R3 \\
13 & Zolera FX 3.34 SC & Yuba & 5 oz/A & R3 \\
14 & Stratego YLD & Bayer & 4 oz/A & R3 \\
15 & Preemptor & Yuba & 5 oz/A & R3 \\
16 & Quadris & Yuba & 6 oz/A & R3 \\
17 & Quadris Top & Yuba & 8 oz/A & R3 \\
18 & Trivapro & Yuba & 13.7 oz/A & R3 \\
19 & Domark 230 & Yuba & 6 oz/A & R3 \\
20 & Viathon & Yuba & 23 oz/A & R3 \\
\bottomrule
\end{tabular}
\end{adjustbox}
\FloatBarrier
\vspace{0.1cm} 
\end{table*}

\begin{table*}[ht!]
\centering
\begin{center}
    \textbf{Fungicide Trial in Year 2019}
\end{center}
\begin{adjustbox}{max width=\textwidth}
\begin{tabular}{cccccc}
\toprule
\textbf{Trt} & \textbf{Product} & \textbf{FRAC} & \textbf{Company} & \textbf{Rate} & \textbf{Timing} \\
\midrule
1 & UTC & --- & All & --- & --- \\
2 & Quadris & 11 & Syngenta & 6 oz/A & R3 \\
3 & Domark 230 & 3 & Gowan & 5 oz/A & R3 \\
4 & Endura & 7 & BASF & 11 oz/A & R3 \\
5 & Priaxor & 7,11 & BASF & 4 oz/A & R3 \\
6 & Preemptor & 3,11 & FMC & 5 oz/A & R3 \\
7 & Miravis Neo & 3,7,11 & Syngenta & 13.7 oz/A & R3 \\
8 & Lucento & 3,7 & FMC & 5 oz/A & R3 \\
9 & Delaro & 3,11 & FMC & 5 oz/A & R3 \\
10 & Stratego YLD & 3,11 & Bayer & 8 oz/A & R3 \\
11 & Aproach Prima & 3,11 & Bayer & 4 oz/A & R3 \\
12 & Quadris Top & 3,11 & DowDuPont & 6.8 oz/A & R3 \\
13 & Restricted & Restricted & Restricted & Restricted & Restricted \\
14 & Restricted & Restricted & Restricted & Restricted & Restricted \\
15 & Quilt Xcel & 3,11 & Syngenta & 8 oz/A & R3 \\
16 & Affiance & Gowan & 14 oz/A & R3 \\
17 & Veltyma & Gowan & 14 oz/A & R3 \\
18 & Revytek & BASF & 7 & R3 \\
19 & Restricted & Restricted & Restricted & Restricted & Restricted \\
20 & Acropolis & AMVAC & 23 & R3 \\
\bottomrule
\end{tabular}
\end{adjustbox}



\begin{center}
    \textbf{Fungicide Trial in Year 2020}
\end{center}
\begin{adjustbox}{max width=\textwidth}
\begin{tabular}{cccccc}
\toprule
\textbf{Trt} & \textbf{Product} & \textbf{FRAC} & \textbf{Company} & \textbf{Rate} & \textbf{Timing} \\
\midrule
1 & UTC & --- & All & --- & --- \\
2 & Quadris & 11 & Syngenta & 6 oz/A & R3 \\
3 & Domark 230 & 3 & Gowan & 5 oz/A & R3 \\
4 & Endura & 7 & BASF & 11 oz/A & R3 \\
5 & Priaxor & 7,11 & BASF & 4 oz/A & R3 \\
6 & Preemptor & 3,11 & FMC & 5 oz/A & R3 \\
7 & Miravis Neo & 3,7,11 & Syngenta & 13.7 oz/A & R3 \\
8 & Lucento & 3,7 & FMC & 5 oz/A & R3 \\
9 & Topguar EQ & 3,11 & FMC & 5 oz/A & R3 \\
10 & Delaro & 3,11 & Bayer & 8 oz/A & R3 \\
11 & Stratego YLD & 3,11 & Bayer & 4 oz/A & R3 \\
12 & Aproach Prima & 3,11 & DowDuPont & 6.8 oz/A & R3 \\
13 & Quadris Top & 3,11 & Syngenta & 8 oz/A & R3 \\
14 & Quilt Xcel & 3,11 & Syngenta & 10.5 oz/A & R3 \\
15 & Affiance & Gowan & 14 oz/A & R3 \\
16 & Veltyma & BASF & 7 & R3 \\
17 & Revytek & BASF & 8 & R3 \\
18 & Restricted & Restricted & Restricted & Restricted & Restricted \\
19 & Acropolis & AMVAC & 23 & R3 \\
20 & UTC & --- & All & --- & --- \\
\bottomrule
\end{tabular}
\end{adjustbox}

\end{table*}

\subsection{Method}

\subsubsection{Yield loss function}
\begin{figure*}[h!]
 \centering
 \includegraphics[width=\textwidth]{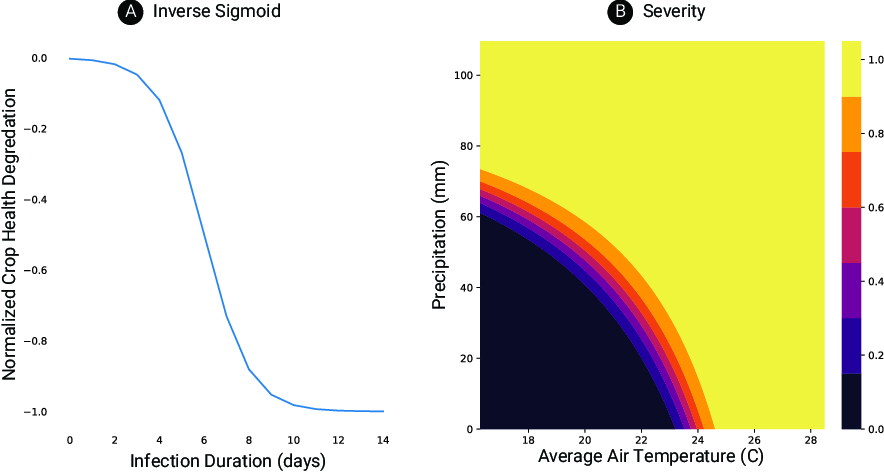}
 \caption{Graphical representation of yield loss factors. (A) Inverse sigmoid function with an x-axis shift of 6 days and y-axis values from $[0,-1]$, represents the yield loss with duration of infection. Once a single crop is infected, as the x-axis increases, the y-axis value decreases, emulating as an infected crop left untreated will eventually lose its harvest over time. (B) Example use case of weather data integration with the environment, depicting a severity plot with average air temperature on the x-axis and precipitation ins the y-axis. The upper right portion of the severity contour plot depicts favorable conditions for biotic stresses to occur with the maximum severity that significantly affects the attainable yield.}
 \label{fig:reward}
\end{figure*}

The yield decay of a crop after being infected and left untreated for a given infection initiation ($G_{inf}$) is modeled by an inverse sigmoid function, which is visualized in Fig.~\ref{fig:reward} (A). The equation computes an individual crop’s yield decay percentage
ranging from $[0, -1.0]$, indicating from $0\%$ to $100\%$ yield loss. For reference, we reproduce the equation where the yield loss ($Y_{loss}$) is a function of the severity of infection ($\mathcal{S}_{inf}$) and the proportion of the attainable yield that is actually realized ($\eta_Y$):

\begin{equation}
   Y_{loss} = (1 - \eta_Y) \times \mathcal{S}_{inf}
   \label{eq:yield_loss2}
\end{equation}
Fig.~\ref{fig:reward} (A), the inverse sigmoid is shifted by \( s = 6 \), moving the initial central point from 0 to 6 days signifying the infection duration by which there will be $50\%$ of the total yield loss. When an individual crop gets infected, the environment tracks the number of infected days if left untreated. As the x-axis increases towards the right of the plot, the degradation of the crop health decreases closer to $-1.0$.

If weather data is used, the user can input an equation to map severity values to contributing factors such as precipitation and average air temperature, as seen in Fig.~\ref{fig:reward} (B). Severity estimates used as an example were adopted from historical studies of Sclerotinia sclerotiorum severity in soybeans~\cite{fall2018case}.
\FloatBarrier

\subsection{Comprehensive Technical Description}
\subsubsection{Reinforcement Learning hyperparameters}
The hyperparameters used for the Double Deep Q-Network (DDQN), Proximal Policy Optimization (PPO), and Trust Region Policy Optimization (TRPO) algorithms are outlined below. Each algorithm's setup was strategically chosen to enhance performance across diverse reinforcement learning scenarios:

\begin{itemize}
  \item \textbf{DDQN}: Features a linearly decaying epsilon strategy, optimizing the trade-off between exploration and exploitation over time.
  \item \textbf{PPO}: Utilizes RMSprop due to its robustness in handling the variability of gradients, crucial for stable training in dynamic environments.
  \item \textbf{TRPO}: Employs minimal hyperparameter tuning, relying on its inherently stable optimization capabilities \cite{schulman2015trust}, especially effective in complex policy spaces.
\end{itemize}

For a detailed breakdown of specific hyperparameters used, refer to Table \ref{tab:hyperparam}.
\begin{table*}[!htb]
\centering
\caption{Hyperparameters for PPO, TRPO, and DDQN algorithms.}
\begin{tabular}{cccc} 
\hline
 & PPO & TRPO & DDQN \\ 
\hline
Optimizer & RMSprop & Adam & LinearDecayEpsilonGreedy \\ 
Learning rate & 0.00075 & & \\
Update interval & 4000 & 5000 & 130 \\
Target update interval & & & 1300 \\ 
Minibatch size & 2250 & & \\ 
Number of epochs (M) & 40 & 10 & \\ 
Clipping parameter $(\varepsilon)$ & 0.284 & & \\
Entropy coefficient & 0.182 & 0 & \\
Gamma $(\Gamma)$ & & 0.995 & 0.99 \\
Replay start size & & & 200000 \\
Decay steps & & & 350000 \\
\hline
\end{tabular}
\label{tab:hyperparam}
\end{table*}

\subsection{Deep RL results} This section provides performance details of different reinforcement algorithms trained on the AgGym environment. 
\subsubsection{Evaluation and Comparison of Algorithms}

We evaluated various reinforcement learning algorithms based on the average reward obtained during a series of 10,000 training episodes. The training was repeated ten times with different random seeds under the growth season setting, and the summarized results are presented in Table~\ref{tab:rewards}. TRPO outperformed other algorithms, achieving the highest expected rewards with the lowest standard deviation, indicating superior consistency and effectiveness.

\subsubsection{Performance Trends Across Episodes}

\begin{figure*}[h]
\centering
\includegraphics[width=15cm,height=10 cm]{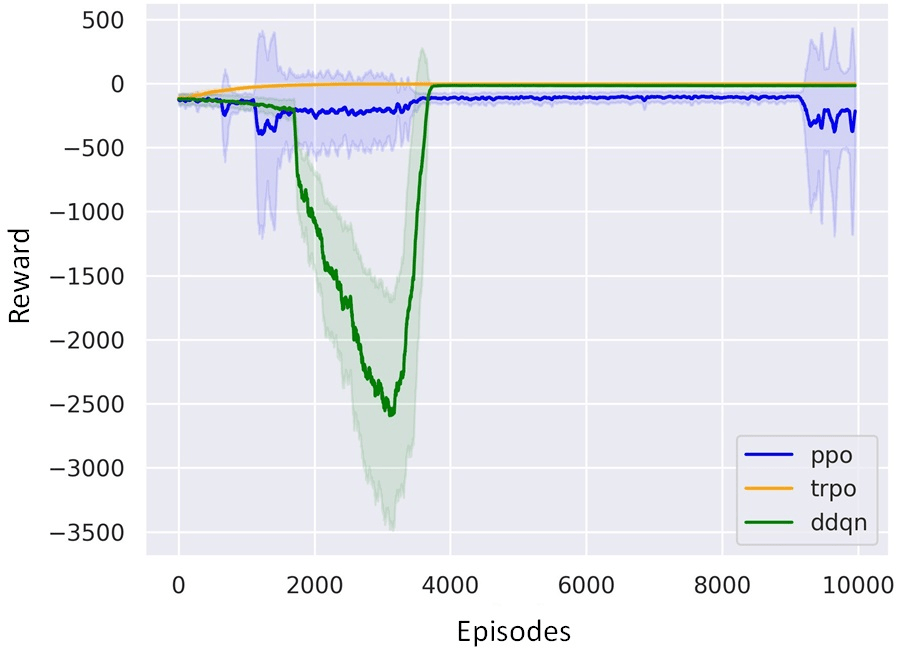}
\caption{Mean cumulated reward of each of the 3 algorithms against the number of episodes.}\label{fig:cumreward}

\end{figure*}

Figure~\ref{fig:cumreward} illustrates the moving average rewards for each algorithm across ten iterations. TRPO converges more rapidly to effective policies than its counterparts. Within about 500 episodes, TRPO approached an optimal policy configuration. Although PPO initially performed well, surpassing DDQN until nearly episode 4000, it was eventually overtaken by DDQN, which adapted a more effective long-term strategy.

\subsubsection{Pesticide Policy and Financial Implications}

\begin{figure*}[ht!]
    \centering
    \subfloat[DDQN]{%
        \includegraphics[width=0.7\textwidth, height=6cm]{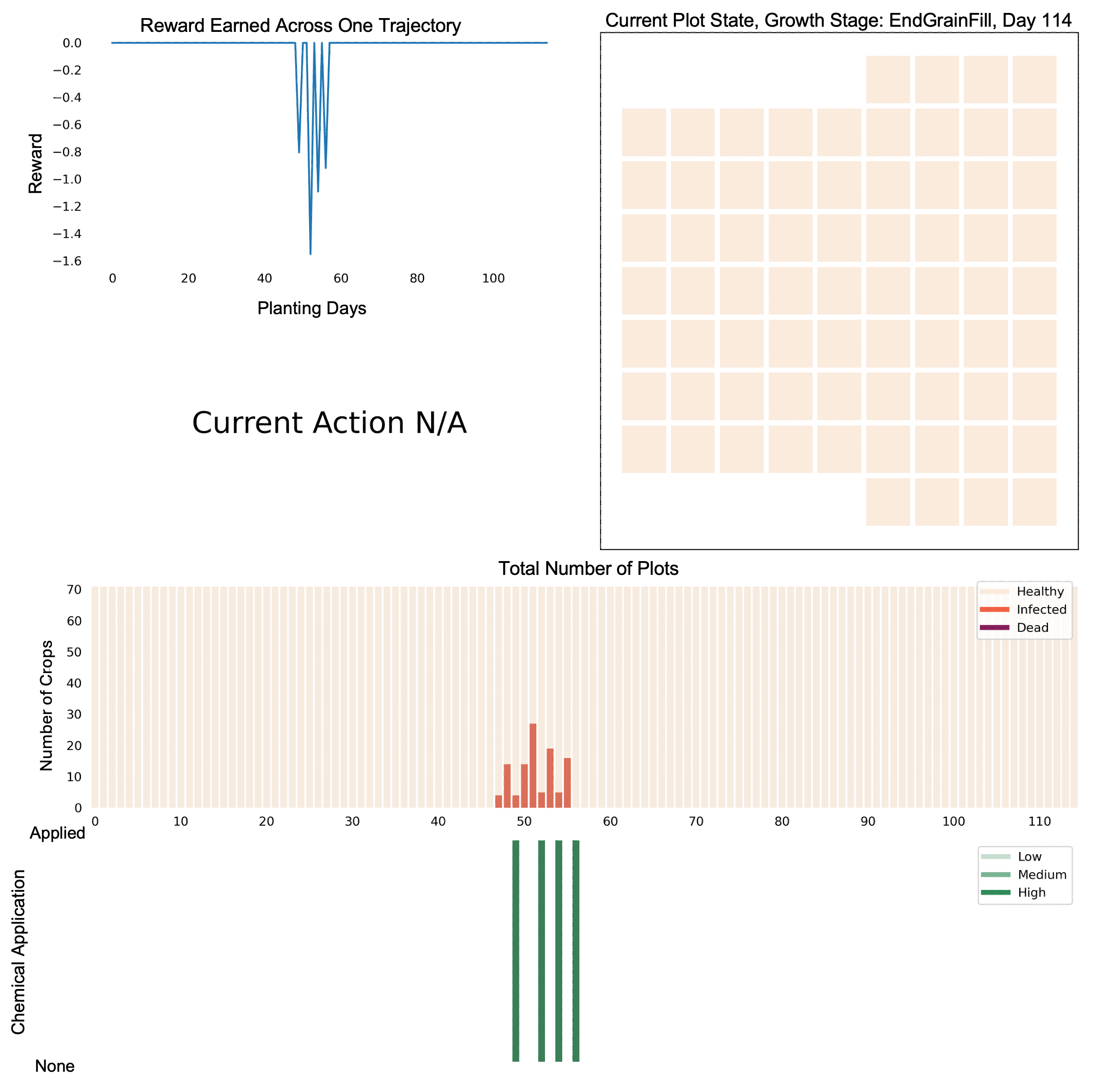}
        \label{fig:ddqn}
    }
    \vfill 
    \subfloat[TRPO]{%
        \includegraphics[width=0.7\textwidth, height=6cm]{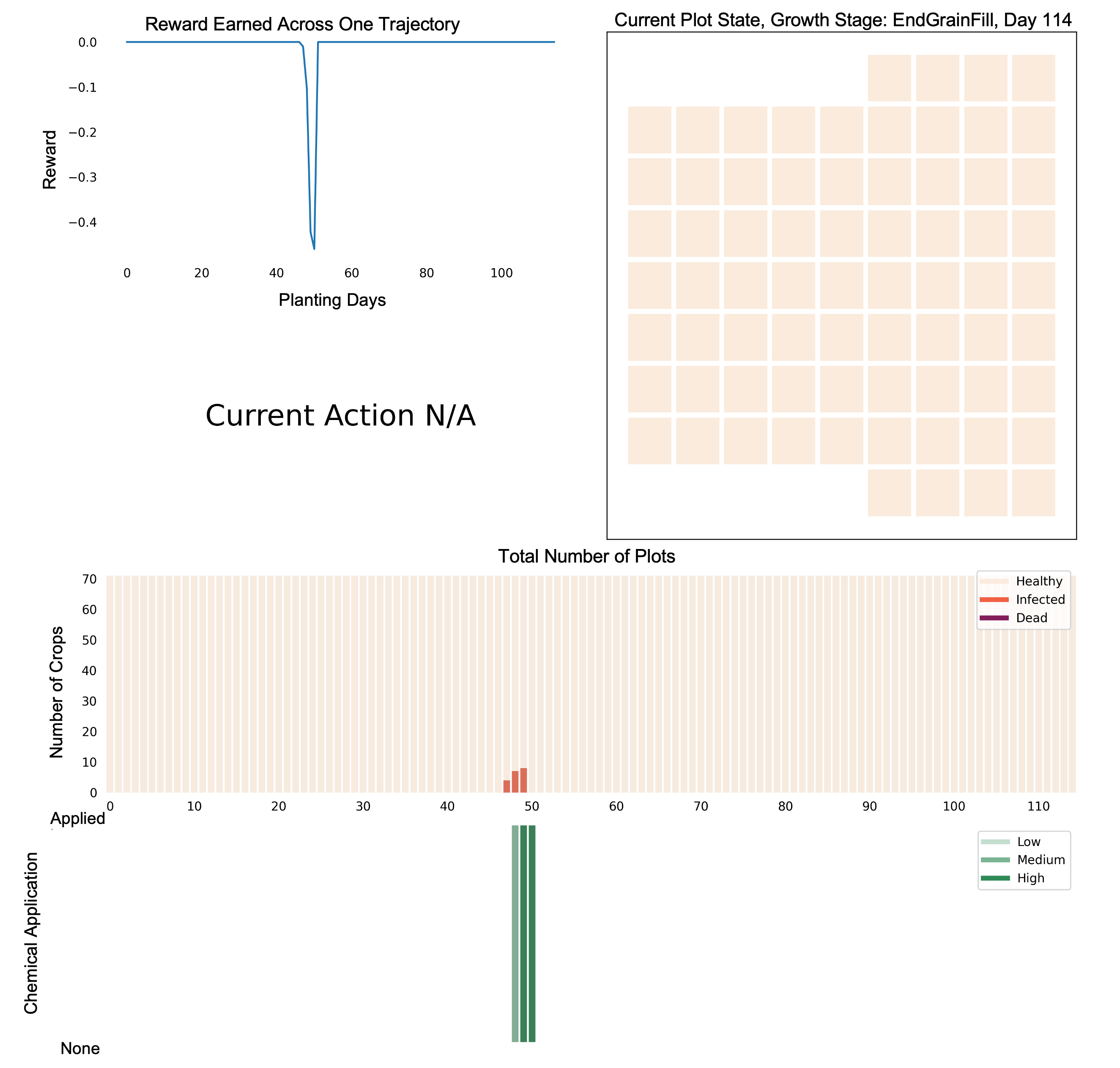}
        \label{fig:trpo}
    }
    \vfill 
    \subfloat[PPO]{%
        \includegraphics[width=0.7\textwidth, height=6cm]{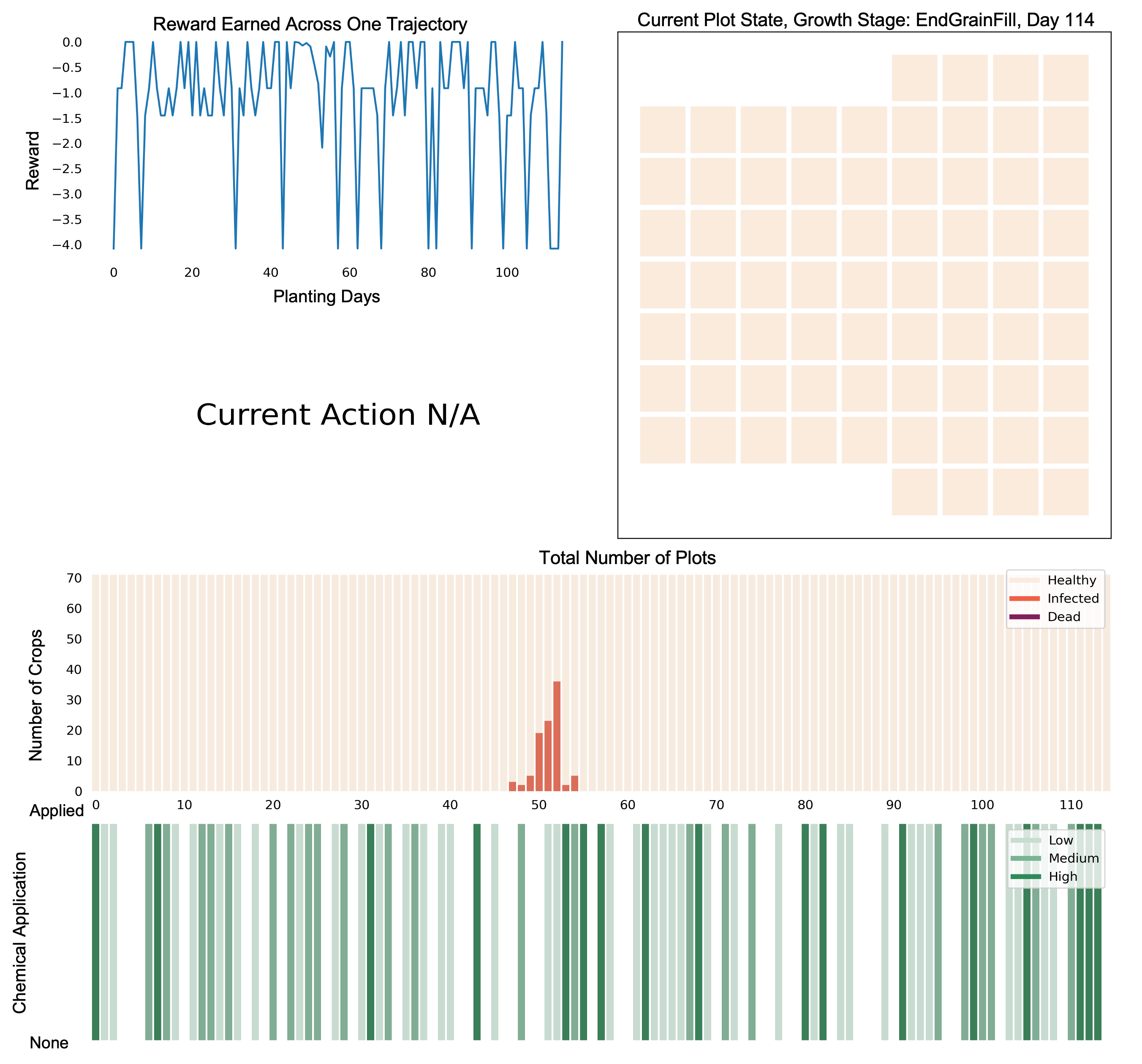}
        \label{fig:ppo}
    }
    \caption{Evaluation of field performance by the (a) DDQN, (b) TRPO, and (c) PPO algorithms on real agricultural fields, illustrating historical actions and responses.}
    \label{fig:performance_breakdown}
\end{figure*}

To further assess the economic implications of each algorithm's strategy, we analyzed their pesticide policies using the most optimal seed from the ten runs. The corresponding field and historical action plots are shown in Figure~\ref{fig:performance_breakdown}. These plots not only depict the financial losses (with rewards approaching zero indicating no loss) but also the frequency of pesticide applications, highlighting the cost-effectiveness and efficiency of the treatments.

Darker bars in the bar plot represent more intensive and costly pesticide applications. Notably, there were no instances of complete crop failure before harvest in any of the simulations, underscoring the effectiveness of the management strategies employed. TRPO and DDQN demonstrated prudent use of pesticides, applying them only as necessary, while PPO incurred higher costs due to frequent applications of varying pesticide qualities, often without adequate justification, resulting in inefficient resource use.

\begin{table*}[ht]
\centering
\captionsetup{width=0.5\textwidth} 
\caption{\centering Average rewards over the 10000 contiguous episodes. $\pm$ denotes standard deviation.}

\label{tab:rewards}
\begin{tabularx}{0.6\textwidth}{Xr}
\toprule
Agent & Mean Reward $\pm$ Std Dev  \\ 
\midrule
DDQN & -170.39 $\pm$ 8.3 \\ 
TRPO & -12.76 $\pm$ 0.94 \\
PPO & -399.45 $\pm$ 35.2 \\
\bottomrule
\end{tabularx}
\end{table*}

\FloatBarrier
\subsection{Environment Design}
This section provides technical details on the operation of AgGym, such as environment structure overview, the input-output pipeline, modularity design of the environment, and reproducibility.
\\

\subsubsection{Environment Structure}
\label{sec:5_1_structure}
The goal of this environment is to enable users to tinker and customize the environment according to their needs. The general environment code structure must be easily customizable to develop new methods, codes, functions, and modules. The code structure is detailed below:

The environment is driven by \texttt{input\_files/} for generating realistic data driven simulation, \texttt{modules/} for customization of different behaviors for threats, agent interactions, and environment components, and finally \texttt{.ini} files to quickly select train and evaluate configurations. For more in-depth details, refer to Sec.~\ref{sec:5_2_1_overview} for \texttt{input\_files/}, Sec.~\ref{sec:5_3_1_modules} for \texttt{modules/}, and Sec.~\ref{sec:5_3_2_ini} for \texttt{.ini}.

\subsubsection{Input-Output Pipeline}
\label{sec:5_2_io}
\textbf{Overview}
\label{sec:5_2_1_overview}

As an end-user for this environment, the necessary input is the configuration files for training or testing. Users can optionally provide shape, weather, and simulation files for higher simulation accuracy since the settings would default to a simplistic assumption for users who want to test the environment. The expected outputs are agent performance visualizations, agent saved weights, performance/cost breakdown, and total revenue.

\textbf{Shapefile}
\label{sec:5_2_2_shapefile}

\begin{figure*}[t!]

\centering
\includegraphics[width=\textwidth]{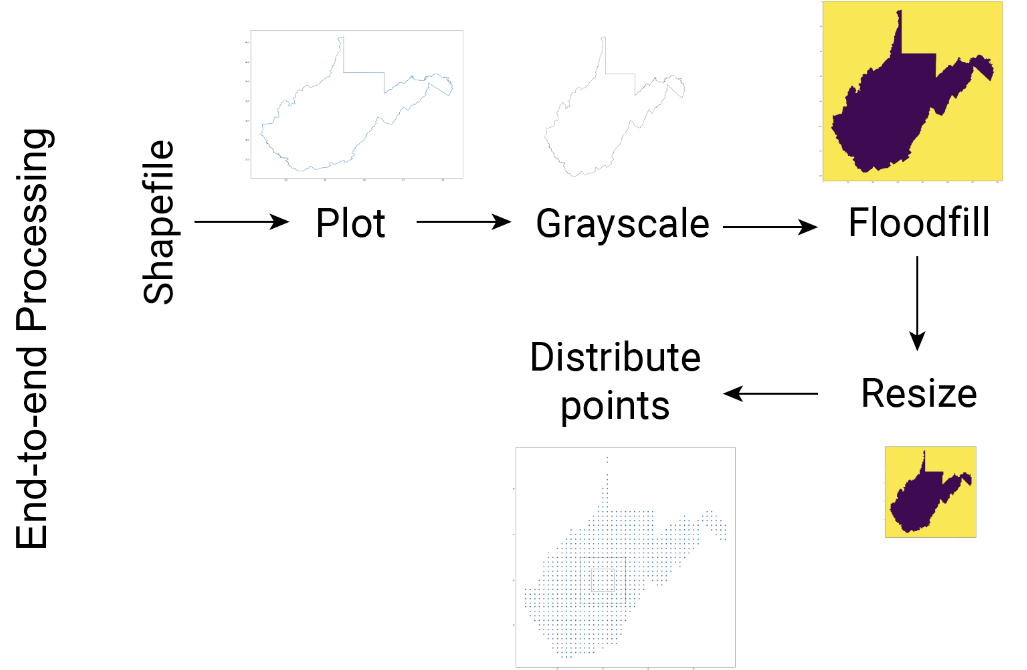}

\caption{End-to-end processing for the input-output pipeline. With the input shapefile, the pipeline will produce a final output of an interactable farm plot for the RL agent. The pipeline includes plotting the x-y coordinates from the shapefile, converting to grayscale to obtain two unique pixel values, \textit{floodfill} to distinguish between inside-outside pixels, and resizing and allocating farm plots based on user input dimensions in List.~\ref{lst:strut}.}
\label{fig:EndtoEnd}
\end{figure*}

A shapefile format stores geospatial information in a vector data format commonly used for geographic information system (GIS) software. Geospatial information can be represented in multiple ways, such as points, lines, and polygons. AgGym environment utilizes polygons to generate a bounded farm plot shaped approximately as the shapefile provided. Figure~\ref{fig:EndtoEnd} illustrates the end-to-end pipeline for processing shapefile to a farm plot. In this case, the state of West Virginia-shaped polygon is plotted on a canvas and converted to a single channel grayscale to prevent variation in pixel values. Once the canvas is converted to grayscale, we utilized a \textit{floodfill} algorithm to fill up the polygon to distinguish between the inside and outside of the polygon. Since the canvas is grayscale, only two distinct pixel values strictly represent inside and outside the polygon. A resizing operation is applied on the canvas, then based on user input dimensions in Sec.~\ref{sec:5_3_modular}, the farm plots will be mapped out in the matrix defined in Sec.~\hyperref[sec:infectionspread]{Infection spread model}.   

\subsubsection{Modularity}
\label{sec:5_3_modular}
\textbf{Modules}
\label{sec:5_3_1_modules}
This environment is defined mostly by functional modules inside \texttt{modules}. The \texttt{modules} consist of three main subfolders, namely \texttt{env}, \texttt{threat}, and \texttt{agent}. Currently, only one environment module inside \texttt{env} is described throughout this paper, but this file structure retains modularity for possible alternate environment versions. For the \texttt{threat} folder, this is where all the open-source threat modules will reside. We are using a simplified threat module, but future contributions from experts can create a diverse library of biotic stress modules that precisely define the dynamics and patterns of spread. Lastly, \texttt{agent} consist of further subfolders with \texttt{action}, \texttt{done}, \texttt{reward}, and \texttt{state}. All four subfolders stores their respective module library which can be expanded in the future. 

\textbf{Initialization File}
\label{sec:5_3_2_ini}

The modularity component relies on users creating modules as mentioned in Sec.~\ref{sec:5_1_structure}, and executed by requesting inside the configuration file as shown in List.~\ref{lst:config}. This listing highlights three sections out of many others in the repository, \textit{input files}, \textit{agent}, \textit{env}, and \textit{cost}. As mentioned in Sec.~\ref{sec:5_1_structure}, providing these files listed under \textit{input files} are optional if the user has specific files to utilize for simulation. 

In \textit{agent} several options defines settings for the RL agent used in the environment. For example, \texttt{action\_type} sets the action type to discrete, continuous, multi-discrete, and so on. \texttt{action} defines the efficacy of the pesticide action and indirectly defines the size of the discrete action space. Finally, \texttt{state} indicates what observation state to expose the agent to train/test with. By default, it is set to crop health; however, if the user wants to add additional auxiliary state information, the user can provide \texttt{+} and the secondary state to use. This will work provided the custom module for the secondary state is written and placed in the correct location as shown in Sec.~\ref{sec:5_1_structure}. 

In the next section, \textit{env} lists multiple components that changes the settings for the AgGym environment. Starting from the top, \texttt{reward} indicates what reward to train the RL agent with. Similarly with the explanation with \texttt{state} in \texttt{agent}, the user can add additional reward functions by invoking the module names, separated by \texttt{+}. The following four components, \texttt{total\_length}, \texttt{total\_width}, \texttt{crop\_length}, \texttt{crop\_width}, defines the total field dimension and individual crop dimensions. In this case, the field is $100 \time 100$, with each crop dimension being $10 \times 10$. That makes a total of a $10 \times 10$ grid matrix if there is no shapefile provided, since the default shape is a square. Finally, the \textit{cost} section is where one can define the financial parameters to take into consideration in the environment. Currently, the environment considers the active parameters that are utilized in the reward function directly such as \texttt{attainable\_yield\_bushel\_per\_acre}, \texttt{revenue\_price\_per\_bushel}, and \texttt{pesticide\_price\_per\_acre}. 
\subsubsection{Reproducibility}
\begin{figure*}[ht!]
\centering
\includegraphics[width=\textwidth]{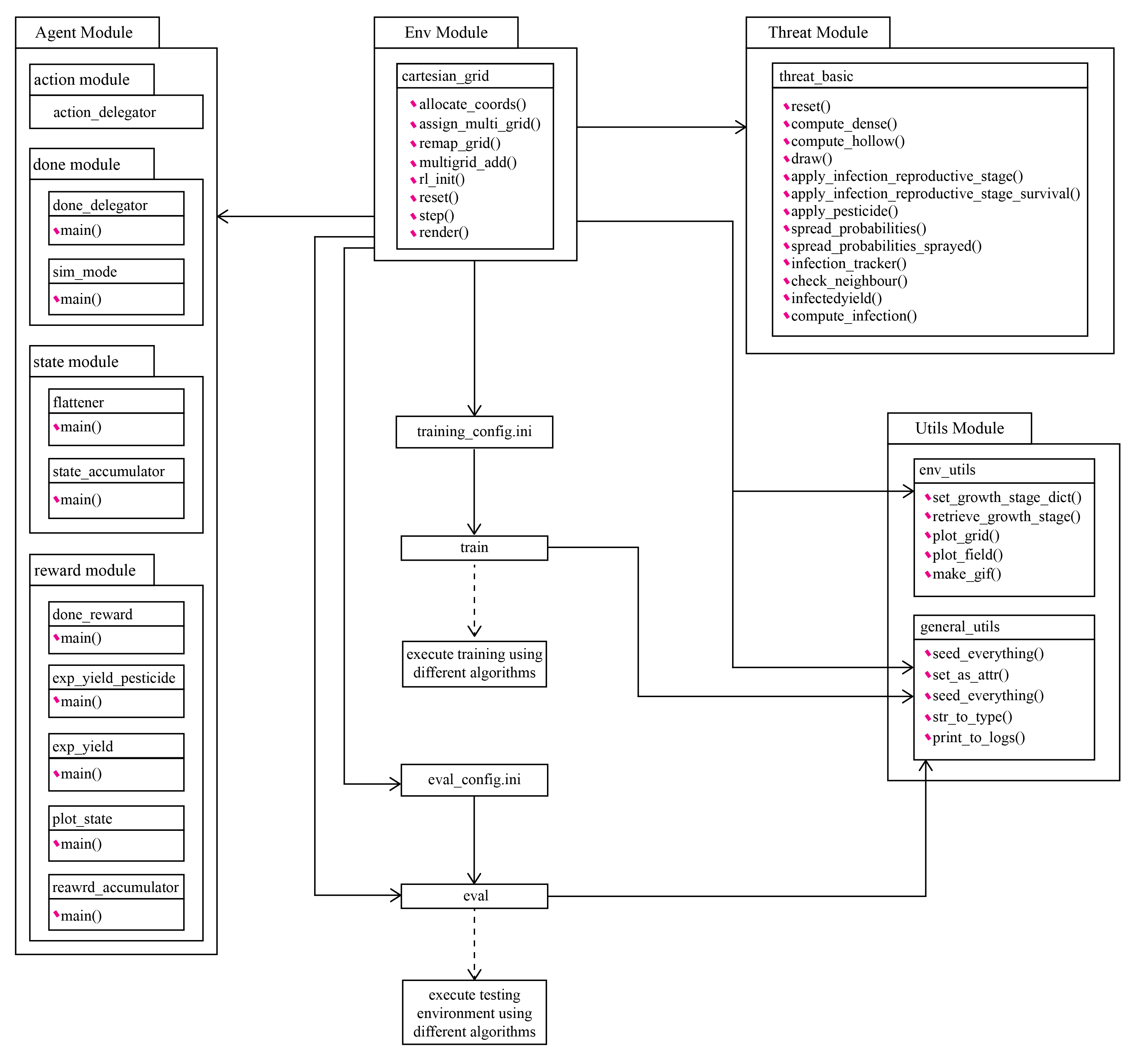}

\caption{UML Diagram illustrating the modular code structure of AgGym simulator. This diagram details the interactions between the \texttt{env\_modules}, \texttt{threat\_modules}, and \texttt{agent\_modules} within the AgGym application. It highlights the flow of data and control across the system, facilitating a better understanding of the high-level architecture and module dependencies.}
\label{fig:uml}
\end{figure*}

AgGym, as an open-sourced project, we strive to make this work as reproducible as possible~\cite{fomel2008guest}. Aside from the detailed description found in this paper, additional descriptions are also in the source code in \href{https://github.com/SCSLabISU/AgGym}{https://github.com/SCSLabISU/AgGym}. All components and functions in AgGym are reproducible since we fixed the random seeds anywhere possible. However, we can only enforce reproducibility up to what we can control. For example, the utilization of non-deterministic operations in deep RL algorithms will cause variations in the final training results~\cite{nagarajan2018impact}. The UML diagram shown in Figure~\ref{fig:uml} provides a visual representation of the modular code structure of the AgGym simulator. It details the interactions between the \texttt{env\_modules}, \texttt{threat\_modules}, and \texttt{agent\_modules}, highlighting the flow of data and control across the system.


x
\begin{figure*}[htbp]
\noindent \begin{minipage}{\linewidth}
    \begin{codeverbatim}[title={Code Structure}, caption={Code Structure}, label={lst:strut}]
    AgGym/
    |
    |-- input_files/
    |   |-- .shp files
    |   |-- .weather files
    |   |-- .simulator files
    |
    |-- modules/
    |   |-- env_modules/
    |   |-- threat_modules/
    |   |-- agent_modules/
    |       |-- action/
    |       |-- done/
    |       |-- reward/
    |       |-- state/
    |   
    |-- train.py
    |-- eval.py
    |-- training.ini
    |-- evaluate.ini
    \end{codeverbatim}
\end{minipage}
\end{figure*}
\label{sec:5_2}
\begin{figure*}[htbp]
\noindent \begin{minipage}{\linewidth}
    \begin{codeverbatim}[title={Configuration File Example}, caption={Configuration File Example}, label={lst:config}]
    
    [input_files]
    shape_file = *.shp
    weather_file = *.csv
    simulator_file = *.csv
    
    [agent]
    action_type = discrete
    action = 0., 0.3, 0.5, 0.9
    state = health
    
    [env]
    reward = r1 + r2
    total_length = 100
    total_width = 100
    crop_length = 10
    crop_width = 10
    multi_level_total_length = 20
    multi_level_total_width = 20
    multi_level_crop_length = 1
    multi_level_crop_width = 4
    
    [cost]
    attainable_yield_bushels_per_acre = 60
    revenue_price_per_bushel = 8
    pesticide_price_per_acre = 17
    
    \end{codeverbatim}
\end{minipage}
\end{figure*}

\end{document}